\documentclass[conference]{IEEEtran}
\usepackage{times}

\usepackage[sort&compress,numbers]{natbib}
\usepackage{multicol}

\usepackage[english]{babel}

\usepackage{amsmath}
\usepackage{amsfonts, amssymb}
\usepackage{amsthm}
\usepackage{bm}
\newcommand{\vect}[1]{\bm{#1}}		\newcommand{\matr}[1]{\bm{#1}}		
\newcommand{\norm}[1]{\left|\left| #1 \right|\right|}
\newcommand{\innerProd}[2]{<#1,#2>}

\newcommand{\pose}{\vect{q}}

\newcommand{\twist}{\dot{\vect{q}}}
\newcommand{\dtwist}{\ddot{\vect{q}}}

\newcommand{\SO}[1]{\mathsf{SO}(#1)}		\newcommand{\SE}[1]{\mathsf{SE}(#1)}		\newcommand{\nR}[1]{\mathbb{R}^{#1}}		

\newtheorem*{remark}{Remark}

\renewcommand{\frame}{\mathcal{F}}																							

\newcommand{\frameW}{\frame_W}																		

		\newcommand{\frameRAbb}{R}																

		\newcommand{\frameHAbb}{H}																

\newcommand{\mfld}[1]{\mathcal{#1}}	        

\newcommand{\mfldC}{\mfld{Q}}

\newcommand{\mfldX}{\mfld{X}}
\newcommand{\dimX}{d_{\mfld{X}}}
\newcommand{\rmpJX}{\matr{J}^\mfldC_\mfldX}

\newcommand{\poseX}{\vect{x}}
\newcommand{\twistX}{\dot{\vect{x}}}

\newcommand{\mfldH}{\mfld{H}}
\newcommand{\dimH}{d_{\mfld{H}}}

\newcommand{\poseR}{\pose_{\frameRAbb}}

\newcommand{\twistR}[1][]{{\twist_{\frameRAbb#1}}}
\newcommand{\dtwistR}{\dtwist_{\frameRAbb}}
\newcommand{\uR}{\vect{u}_{\frameRAbb}}

\newcommand{\rmpF}{\vect{f}}
\newcommand{\rmpA}{\matr{A}}
\newcommand{\rmpJ}{\matr{J}}

\newcommand{\rmpPot}{\Phi}
\newcommand{\rmpdPot}{\vect{\nabla}\rmpPot}
\newcommand{\rmpdPotHuman}{\rmpdPot_{des}^\mfldH}

\newcommand{\rmpScalei}{\alpha}

\newcommand{\rmpScaleDelta}{\alpha_{\Delta}}

\newcommand{\rmpScaleAngle}{\gamma}

\newcommand{\uH}{\vect{u}_{\frameHAbb}}
\newcommand{\rH}{r^\mfldH}
\newcommand{\poseH}{\vect{h}}

\newcommand{\poseHR}{\vect{h}_{\frameRAbb}}
\newcommand{\poseHROpt}[1]{\vect{h}_{\frameRAbb,#1}^\star}
\newcommand{\twistHR}{\dot{\vect{h}}_{\frameRAbb}}

\newcommand{\likelihoodPost}[2]{P\left(\left.#1\right|#2\right)}
\newcommand{\likelihood}[1]{P\left(#1\right)}
\newcommand{\likelihoodTotShort}{P_i}

\newcommand{\eigVal}{\lambda}
\newcommand{\eigVec}{\vect{v}}

\usepackage{todonotes}
\let\labelindent\relax
\usepackage[inline]{enumitem}
\usepackage{siunitx}

\usepackage[linesnumbered,ruled,vlined]{algorithm2e}

\usepackage{acro}
\DeclareAcronym{RMP}{ 
    short = {RMP}, 
    long  = {Riemannian Motion Policy},
    long-plural-form  = {Riemannian Motion Policies},
}

\DeclareAcronym{RL}{
    short = {RL},
    long = {Reinforcement Learning},
}

\DeclareAcronym{DoF}{
    short = {DoF},
    long = {degrees of freedom},
}

\DeclareAcronym{WLOG}{
    short = {WLOG},
    long = {without loss of generality}
}

\DeclareAcronym{POMDP}{
    short = {POMDP},
    long = {Partially Observable Markov Decision Process}
}

\DeclareAcronym{HRI}{
    short = {HRI},
    long = {Human-Robot Interaction}
}

\usepackage{tikz}
\usetikzlibrary{shapes,arrows,calc,fit}
\tikzstyle{node} = [rectangle, text centered, draw=black, rounded corners, align=center]

\usepackage{multicol}
\usepackage{multirow}
\usepackage{subfig}
\usepackage[bookmarks=true,urlcolor=cyan,colorlinks=false]{hyperref}
\usepackage{cleveref}

\usepackage[normalem]{ulem}
\newcommand{\add}[1]{#1}
\newcommand{\remove}[1]{}

\begin{document}
\newcommand{\Title}{Task Adaptation in Industrial Human-Robot Interaction: Leveraging Riemannian Motion Policies}

\title{\Title}

\author{
    \authorblockN{Mike~Allenspach,
    Michael~Pantic,
    Rik~Girod, 
    Lionel~Ott and
    Roland~Siegwart}
    \authorblockA{All authors are with the Autonomous Systems Lab, ETH Zurich, 8092 Zurich, Switzerland.\\
{\tt\small $\{$\href{mailto:amike@ethz.ch}{amike}, \href{mailto:mpantic@ethz.ch}{mpantic}, \href{mailto:brik@ethz.ch}{brik}, \href{mailto:lott@ethz.ch}{lott}, \href{mailto:rsiegwart@ethz.ch}{rsiegwart}$\}$@ethz.ch} }}

\maketitle

\begin{abstract}
In real-world industrial environments, modern robots often rely on human operators for crucial decision-making and mission synthesis from individual tasks. 
Effective and safe collaboration between humans and robots requires systems that can adjust their motion based on human intentions, enabling dynamic task planning and adaptation.
Addressing the needs of industrial applications, we propose a motion control framework that 
\begin{enumerate*}[label=(\roman*)]
\item removes the need for manual control of the robot's movement;
\item facilitates the formulation and combination of complex tasks; and
\item allows the seamless integration of human intent recognition and robot motion planning.
\end{enumerate*}
For this purpose, we leverage a modular and purely reactive approach for task parametrization and motion generation, embodied by Riemannian Motion Policies.
The effectiveness of our method is demonstrated, evaluated, and compared to a representative state-of-the-art approach in experimental scenarios inspired by realistic industrial \acl{HRI} settings.

\end{abstract} 
\section{Introduction}
\label{sec:intro}
With continuous research in artificial intelligence and automation, recent works have demonstrated the potential of autonomous mobile robots in various industrial applications, including infrastructure inspection~\cite{2017-LatMil,2024-PfaBodCro}, construction~\cite{2022-EglGasKer,2023-hilti}, and physical interaction~\cite{2022-OllTogSua,2023-RizChuGaw}.
However, modern robots often still lack the decisional autonomy required for deployment in real-world scenarios with uncertain operating conditions.
This encompasses challenges such as disturbed or unknown task hierarchies (e.g., determining which object to pick up first) and addressing unexpected situations (e.g., handling failed measurements at inspection targets).

Therefore, operational safety is typically guaranteed by a human operator that \add{provides} situational awareness, reasoning, and problem-solving skills. Apart from safety considerations, combining the complementary capabilities of humans and robotic systems \add{is beneficial to} the overall task performance and completion time~\cite{2021-YanZhuChe}.

Augmenting a (partially) autonomous system with real-time human inputs can be realized in different ways, ranging from low-level manual control of the robot's movements~\cite{2020-CoeSinKon,2022-AllLawTog} to high-level control by indicating the current mission objective~\cite{2017-StoCheTil}.
Although the former offers complete motion control, the latter tends to be less strenuous for the operator during repetitive and precision-demanding tasks~\cite{2021-SelCogNik}.
Additionally, recent investigations in~\cite{2019-YouMilBi,2022-AllLawTog} indicate that manual control of robots is inefficient, especially for systems involving a high number of controllable \acp{DoF}. 

\begin{figure}
    \centering
    \includegraphics[width=\linewidth]{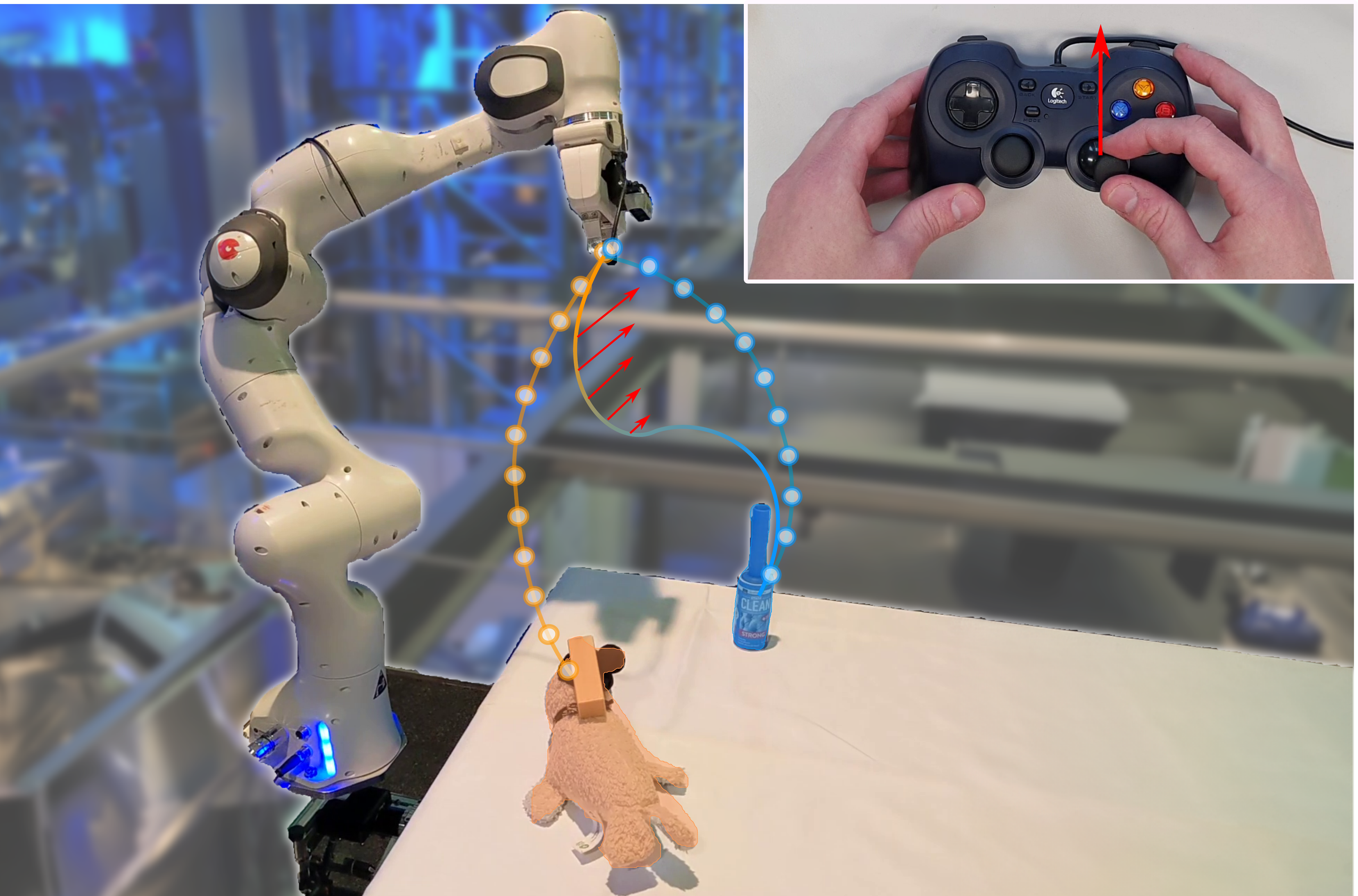}
    \caption{\add{A motivating application for this work is obstacle grasping with unknown task hierarchy. The robot possesses motion policies for handling both objects but initially prioritizes the orange one for pickup. However, the human operator envisions a different order and issues corrective commands via a gamepad to convey their desired direction of motion. 
    The task is dynamically adapted to pick up the blue object instead.}
}    
    \label{fig:teaser}
\end{figure}

As such, our objective is to develop an exclusively autonomous motion control framework tailored to industrial applications. 
The goal is to reduce the operator's workload by eliminating manual robot motion generation while still preserving control authority.
Inspired by recent advances in robot motion generation, we use \acp{RMP}~\cite{2018-RatIssKap} \add{as a} purely local reactive motion generation framework that provides a simple yet powerful, mathematically and geometrically consistent~\cite{2018-CheMukIss} task specification.
A motivating example that showcases the desired functionality of the framework is illustrated in \Cref{fig:teaser}.

\subsection{Related Work}
The alignment of robot motions with human intentions has gained increasing interest in the research community, particularly in the context of safe deployment of autonomous systems in industrial settings~\cite{2021-SelCogNik,2022-OllTogSua,2021-YanZhuChe}.
Throughout the extensive literature, a consistent assumption \add{is} that the human operator has a particular reward function in mind, representing their preferences for mission execution.
In this context, it is crucial to note that the scope of industrial \ac{HRI} significantly differs from more human-centered research fields, such as household or rehabilitation robotics.
There, the emphasis lies in optimizing execution details for human comfort~\cite{2018-LosMcdBatt}, considering factors like distance to humans and obstacles~\cite{2021-LosBajOma,2020-LosOmaMar}, based on expert demonstrations of a specific task~\cite{2016-FinLevAbb,2016-HadDraAbb,2020-RavPolChe}.
Conversely, industrial \ac{HRI} focuses on simplifying ad-hoc dynamic task planning and adaptability. 
While the constituting tasks of industrial applications are commonly predefined~\cite{2018-RatIssKap}, their specific order may be unknown or evolve in real-time based on changing operational requirements.
Two sets of issues are tackled by inferring the unknown reward from human actions; compliance of the robot's motion to the human input~\cite{2022-FraPedBes}, and active reconfiguration of the task hierarchy.

The unknown human reward function is commonly formulated as a linear combination of individual task features~\cite{2008-ZieMaaBag}.
As such, inferring the reward function simplifies to determining the appropriate relative weights.
Corresponding data-driven methods based on Neural Networks~\cite{2016-WulOndPos,2015-WulOndPos,2020-CheSunLiu}, as well as offline ~\cite{2022-AshJhaPil} and online \ac{RL}~\cite{2017-RhiKit,2018-RedDraLev} have been derived.
Despite their success in accomplishing complex tasks, such data-driven approaches generally require a (potentially large) amount of expert demonstrations and are highly sensitive to changes in the mission setup, robotic system, and environment.

Alternatively, model-based methods remove the need for prior data collection, instead only observing the real-time user input. 
Initial works concentrated on grasping scenarios, where task features corresponded to the objects to be picked up~\cite{2013-DraSri,2017-JavAdmPel,2015-JavBagSri}.
Based on the user input history, the likelihood of each object is continuously updated until the system is confident about the next object to be picked. 
In addition to being rather slow, the proposed methods often require approximations and simplification for tractability.
Instead, \cite{2019-KhoBil} \add{uses} dynamical systems for task adaptation, enabling the parametrization of generic tasks and guaranteeing stability.

Many of the existing approaches have limitations regarding their applicability for generic industrial applications.
Firstly, the manual control they typically employ is inefficient~\cite{2019-YouMilBi,2022-AllLawTog}, inspiring an \textit{exclusively autonomous} motion control framework.
Secondly, the presented methodologies lack the ability to infer independent task features, emphasizing the need for a more \textit{modular} approach.
Thirdly, formulating \add{all} task features \add{in a single coordinate frame or manifold} \add{is often} challenging, motivating a \textit{purely local} task parametrization.

\subsection{Contributions}
To address the identified limitations of existing task adaptation approaches, we propose an \textit{exclusively autonomous} robot motion control framework that relies on \textit{highly modular} and \textit{purely local} reactive motion policies. A state-of-the-art implementation is provided by \acp{RMP}~\cite{2018-RatIssKap}, which offer high-computational efficiency~\cite{2013-PanMeiBah}, inherent system agnosticism and robustness~\cite{2023-LanPanBah}, as well as a well-defined geometric-mathematical structure~\cite{2018-CheMukIss}.

As such, the main contributions of this work are:
\begin{itemize}
    \item 
    Design of an exclusively autonomous motion control framework for dynamic task planning and adaptation;
    \item Demonstration and evaluation of the proposed approach's effectiveness;\item Discussion and comparison of the proposed approach to a \add{representative} state-of-the-art implementation~\cite{2019-KhoBil}.
\end{itemize}
In the following sections, we present the main building blocks of our system (\Cref{sec:setup}), describe how task adaption is realized through the inferred re-weighting of individual motion policies (\Cref{sec:method}), introduce our experimental setup (\Cref{sec:exp_setup}), and perform a quantitative evaluation of the proposed approach (\Cref{sec:results}).

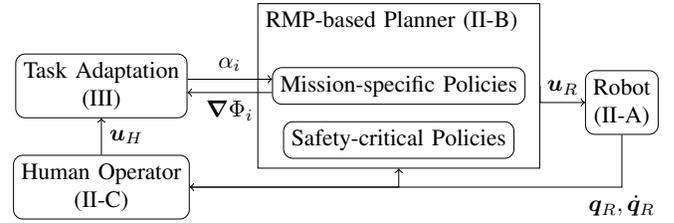
\begin{figure}
    \centering
        \resizebox{\linewidth}{!}{
                \begin{tikzpicture}
\def\hnodesep{0.55cm}
            \def\vnodesep{0.25cm}
            \def\textdpt{2cm}

\node [node] (human) {Human Operator~\\(\ref{ssec:human})};

\node [node, above = \hnodesep  of human] (reward) {Task Adaptation~\\(\ref{sec:method})};

\node [draw, right = \hnodesep+0.5cm of reward, text depth=\textdpt, text width=4cm] (rmp) {\ac{RMP}-based Planner~(\ref{ssec:rmp_planner})};

            \node [node] (arb_rmp) at (rmp) {Mission-specific Policies};
            \node [below = \vnodesep of arb_rmp, fit=(rmp)] (aux1) {};
            \node [node, below = \vnodesep of arb_rmp] (auto_rmp) {Safety-critical Policies};

\node [node, right = \hnodesep of aux1] (robot) {Robot~\\(\ref{ssec:robot})};

\draw[->] (robot.south) |- node[midway,below]{$\poseR,\twistR$} (human.east);
\draw[->] (human.east) -| (rmp.south);

\draw[->] (human.north) -- node[midway,right]{$\uH$} (reward.south);
            \draw[->] ($(reward.east)+(0,0.1)$) -- node[midway,above]{$\rmpScalei_i$}($(arb_rmp.west)+(0,0.1)$);
            \draw[<-] ($(reward.east)+(0,-0.1)$) -- node[midway,below]{$\rmpdPot_i$} ($(arb_rmp.west)+(0,-0.1)$);

\draw[->] (rmp.east)  |- node[near end, above] {$\uR$} (robot.west);

        \end{tikzpicture}         }
    \caption{Motion control framework for dynamic task adaptation: The human operator observes the robot's motion $\twistR$ and issues corrective commands $\uH$, initiating the adaptation of the current task and corresponding adjustment of the planned motion in the \ac{RMP} planner.}
    \label{fig:system}
\end{figure}

\section{System Overview}
\label{sec:setup}
In this section, we introduce the main elements of the system and provide a succinct problem definition.
The main components are depicted in \Cref{fig:system}, comprising a i) \textit{robot} whose motion is guided by a ii) \textit{\ac{RMP}-based planner}, and a iii) \textit{human operator} whose input influences certain elements of the planner.

\subsection{Robot}
\label{ssec:robot}
\add{Our} method is robot-agnostic, however in this study we use a robotic arm with full pose controllability at the end-effector. Therefore, \ac{WLOG}, we consider the robot's end-effector configuration space manifold $\mfldC$ to be $\SE{3}$, where pose and twist are denoted by $\poseR\in\nR{6}$ and $\twistR\in\nR{6}$, respectively.
We treat the closed-loop dynamics as a double integrator $\dtwistR = \uR$, where the virtual input $\uR\in\nR{6}$ is calculated by the motion planner and represents the instantaneous reference acceleration to be executed by the robot.

\begin{remark}
    \add{The choice of generalized coordinates and configuration space inherently arises from the type of robot and control scheme.}
    \add{However, }the \ac{RMP} motion planner facilitates straightforward transformations between different spatial manifolds (see \Cref{ssec:rmp_planner}) \add{and handles joint or actuator limits naturally~\cite{2018-RatIssKap}}.
    Consequently, the system model is agnostic to specific robot configurations, including for example underactuated, joint- \add{or pose-}controlled platforms.
\end{remark}

\subsection{\acf{RMP}-based Planner}
\label{ssec:rmp_planner}
\acp{RMP} decompose complex robot motion generation problems into several independent policies. Each policy operates within a designated manifold, where the associated problem is easiest to solve. Below is a concise overview of key attributes and operators as outlined in~\cite{2018-RatIssKap,2018-CheMukIss}.
\subsubsection{Policy Definition}
The essential building block, a motion policy, is defined by two elements: a state-dependent acceleration $\rmpF$ and an associated Riemannian metric $\rmpA$, both defined on a dedicated manifold $\mfldX$.
Formally, we define the manifold $\mfldX$ of dimensionality $\dimX$ with generalized coordinates $\poseX:\mfldX\rightarrow\nR{\dimX}$ and $\twistX\in\nR{\dimX}$.
A policy on this manifold is then denoted by the tuple $\left(\rmpF^\mfldX,\rmpA^\mfldX\right)$, with the state-dependent acceleration $\rmpF^\mfldX(\poseX,\twistX):\nR{\dimX}\times\nR{\dimX}\rightarrow\nR{\dimX}$ and the smoothly varying, positive semi-definite Riemannian metric $\rmpA^\mfldX(\poseX,\twistX):\nR{\dimX}\times\nR{\dimX}\rightarrow\nR{\dimX\times\dimX}_{\geq0}$.

While the specific form of $\rmpF^\mfldX$ varies between policies, stability considerations necessitate a structured expression:
\begin{align}
    \rmpF^\mfldX(\poseX,\twistX) &=
    -\rmpdPot^\mfldX(\poseX) - \beta\twistX,
    \label{eq:rmp_f}
\end{align}
where $\beta\in\nR{}_{\geq 0}$ is a damping coefficient.
This formulation guarantees asymptotically stable minimization of the underlying smooth potential function $\rmpPot^\mfldX:\nR{\dimX}\rightarrow\nR{}_{\geq0}$\add{~\cite{2018-RatIssKap}}.

\begin{remark}
    Despite the seemingly restrictive nature of this formulation, $\rmpPot$ enables the expression of intricate motions~\cite{2020-RanLiRav}. 
\end{remark}

The form of the metric $\rmpA^\mfldX$ is not standardized. Rather, it permits the weighting of individual policies relative to others, directionally or axis-wise. 

\begin{remark}
    As outlined in~\cite{2018-RatIssKap}, directional weighting can, for example, improve obstacle avoidance. Compared to the superposition of collision controllers or potential fields, considering directionality ensures that the robot smoothly navigates around obstacles, preventing sudden stops and unintuitive behaviors.
\end{remark}

\subsubsection{Manifold Mapping}
To locally map policies from an arbitrary manifold $\mfldX$ to the configuration space $\mfldC$, we define the map $\vect{\psi}^\mfldC_\mfldX:\nR{6}\rightarrow\nR{\dimX}$ and the corresponding analytic Jacobian $\rmpJX\in\nR{\dimX\times6}$ as:
\begin{align}
    \poseX &= \vect{\psi}^\mfldC_\mfldX(\pose) &
\rmpJX &= 
    \left. \frac{\partial\vect{\psi}}{\partial\pose}\right|_{\poseR}
    \label{eq:rmp_map}
\end{align}
The original policy can then be transformed into an equivalent configuration space policy using the \textit{pull-back} operator \add{defined in~\cite{2018-RatIssKap}}:
\begin{small}
\begin{align}
    \left(\rmpF^\mfldC, \rmpA^\mfldC\right) &=  pull_{\vect{\psi}}\left(\left(\rmpF^\mfldX,\rmpA^\mfldX\right)\right)    \label{eq:rmp_pull} \\
    &:= \left(\left({\rmpJX}^\top\rmpA^\mfldX\rmpJX\right)^\dagger{\rmpJX}^\top\rmpA^\mfldX\rmpF^\mfldX, 
    {\rmpJX}^\top\rmpA^\mfldX\rmpJX\right)\nonumber,
\end{align}
\end{small}
\add{where $^\dagger$ is the matrix pseudo-inverse.}

\subsubsection{Policy Combination}
Given a set of policies $\left(\rmpF_i^\mfldC,\rmpA_i^\mfldC\right),\ i=1,\dots,N$ in configuration space, we define the \ac{RMP} addition operator $\boxplus$ to compute the overall motion policy $\left(\rmpF_{tot}^\mfldC,\rmpA_{tot}^\mfldC\right)$ as a metric-weighted average\add{~\cite{2018-RatIssKap}}:
\begin{align}
    \left(\rmpF_{tot}^\mfldC,\rmpA_{tot}^\mfldC\right)
    &= \underset{i}{\boxplus} \left(\rmpF_{i}^\mfldC,\rmpA_{i}^\mfldC\right) 
        \label{eq:rmp_sum}
\\
    &:=
    \left(
    \left(\sum_i\rmpA_i^\mfldC\right)^\dagger \sum_i\rmpA_i^\mfldC\rmpF_i^\mfldC,
    \sum_i\rmpA_i^\mfldC
    \right).\nonumber
\end{align}

\subsection{Human Operator}
\label{ssec:human}
We model the human operator input as a vector on the manifold of admissible human inputs $\mfldH$. Formally, $\mfldH$ is of dimensionality $\dimH$ and admits generalized coordinates $\poseH:\mfldH\rightarrow\nR{\dimH}$. 
\add{Note that we permit $\mfldH$ to have arbitrary topology and dimensionality, maximizing the utilization of application geometry for an intuitive user interface (see \Cref{sec:exp_setup}).}

The human input signal $\uH\in\nR{\dimH}$ serves as a corrective action aimed at achieving the optimal robot motion.
In other words, the operator observes the robot state $\poseHR=\vect{\psi}_\mfldH^\mfldC(\poseR)\in\nR{\dimH}$ and motion $\twistHR=\rmpJ_\mfldH^\mfldC\twistR\in\nR{\dimH}$ and indicates the direction and magnitude adjustments required to achieve their desired task.

\subsection{Problem Formulation}
\label{ssec:problem}
\subsubsection{Task Parametrization}
\label{sssec:task_parametrization}
Generic industrial applications can be decomposed into multiple tasks, each defined by a set of inherent features to optimize.
In this work, we parametrize tasks as a linear combination of \acp{RMP}, each dedicated to optimizing one specific task feature.
Mathematically, this is expressed by modeling all features $i=1,\dots,N$ as potential functions $\rmpPot_i^{\mfldX_i}$ in a designated manifold $\mfldX_i$.
Optimizing these task features using \acp{RMP} then becomes straightforward through the definition of a set of policies $\left(\rmpF_i^{\mfldX_i},\rmpA_i^{\mfldX_i}\right)$ with accelerations following \Cref{eq:rmp_f}.
We categorize policies as either
\begin{enumerate*}[label=(\roman*)]
    \item \textit{safety}-critical (e.g., collision avoidance, workspace limits, joint constraints), with metrics $\rmpA_i^{\mfldX_i}$ designed to dominate all other policies in critical conditions (\add{see~\cite{2018-RatIssKap,2018-CheMukIss}}), or
    \item \textit{mission}-specific (e.g., targets to inspect, objects to pick up), allowing for dynamic adjustments in priorities as the task adapts by gradual modifications of the relevant metrics $\rmpA_i^{\mfldX_i}$.
\end{enumerate*}

\begin{remark}
    While motion policies must be tailored to specific mission conditions, simple attractor and repulsor \acp{RMP} often suffice for describing even complex behaviors~\cite{2023-LanPanBah}.
    These policies, characterized by well-defined potential functions, can be repurposed by simply adjusting the extremum point, e.g. based on task feature and obstacle locations.
    Independent of the reuse of policies, the proposed adaptation methodology (see \Cref{sec:method}) remains application-agnostic, requiring \acp{RMP} to adhere only to the structured expression in \Cref{eq:rmp_f}.
\end{remark}

For dynamic task adaptation, we allow the humans to control the system by indirectly scaling the metric of the $N_{mis}$ mission-specific policies through their inputs, while the $N_{saf}=N-N_{mis}$ safety-critical ones remain unaffected.
Restricting human intervention solely to mission aspects not only simplifies task adaptation but also prevents accidental overwriting of safety features~\cite{2017-LasSon}.
Considering this, the robot motion reference $\uR$ is given in configuration space $\mfldC$ as:
\begin{align}
    \left(\uR,\cdot\right) = 
    \overset{N_{mis}}{\underset{i=1}{\boxplus}}\left(\rmpF_i^\mfldC, \rmpScalei_i\rmpA_i^\mfldC\right)
    + \overset{N}{\underset{i=N_{mis}+1}{\boxplus}} \left(\rmpF_i^\mfldC, \rmpA_i^\mfldC\right),
    \label{eq:task_motion}
\end{align}
where the scaling values $\alpha_i\in[0,1]$ specify the human intended task and are inferred based on their corrective action $\uH$ \add{by the proposed task adaptation} (see \Cref{sec:method}).
Here, all policies are pulled to $\mfldC$ according to \Cref{eq:rmp_pull}, evaluated at the robot's current pose $\poseR$ and twist $\twistR$, and finally aggregated following \Cref{eq:rmp_sum}.

\begin{remark}
In line with the fully autonomous setup motivated in \Cref{sec:intro}, it is crucial to note the absence of direct manual control over robot motion by the user. Instead, user influence is indirect via metric scale adjustments, eliminating the need for designing suitable arbitration laws between human and planner~\cite{2013-DraSri}. Additionally, robot motion is more deterministic, driven by globally defined and provably stable motion policies.
\end{remark}

\subsubsection{Human Reward}
Given our choice of task parametrization (see \Cref{sssec:task_parametrization}), each motion policy in \Cref{eq:task_motion} fundamentally optimizes the associated task feature.
Therefore, scaling the \add{former's} metrics by $\rmpScalei_i$ is equivalent to adjusting the latter's relative importance, enabling the modeling of the human objective $\rH:\nR{\dimH}\rightarrow\nR{}_{\leq0}$ in the input manifold $\mfldH$ as a linear combination of all task feature potentials~\cite{2008-ZieMaaBag}:
\begin{align}
    \rH(\poseHR) &= -\sum_{i=1}^{N_{mis}}\rmpScalei_i\rmpPot_i^\mfldH(\poseHR) =: -\rmpPot_{des}^\mfldH(\poseHR)
    \label{eq:task_reward}
\end{align}
Following \Cref{ssec:human}, the operator issues corrective actions $\uH$ to optimize the robot configuration with respect to this reward function.
In practical terms, this involves achieving a robot velocity $\twistHR$ aligned with the gradient of $\rH$, i.e., along the negative gradient of the desired potential $-\rmpdPotHuman(\poseHR)$. 
This implies that the human effectively indicates the desired potential function gradient, which is a crucial postulate for the task adaptation discussed in \Cref{sec:method}.

\begin{remark}
Users could adjust the policy scaling directly to convey their desired objective, but such explicit communication is often ineffective~\cite{2008-GreBilChe,2003-GooJr}. This motivates the adoption of a more implicit form of information exchange, like the natural indication of desired robot motion pursued here.
\end{remark}

\section{Task Adaptation}
\label{sec:method}
\begin{figure}
    \centering
    \includegraphics[width=0.75\linewidth]{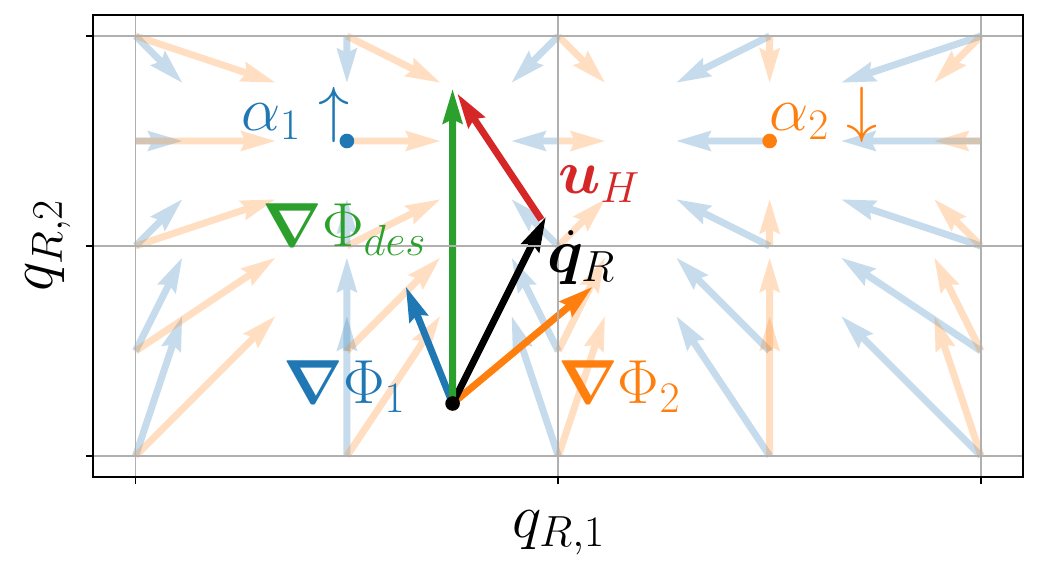}
    \caption{
    A 2D task adaptation example, resembling a top-down view of the scene in \Cref{fig:teaser}:    
    The human-desired direction of motion $\bm{\nabla}\Phi_{des}$ aligns better with the first policy $\bm{\nabla}\Phi_{1}$. The task adaptation prioritizes it $\rmpScalei_1\uparrow$ over the second one $\rmpScalei_2\downarrow$.}
    \label{fig:2D_example}
\end{figure}

The individual mission-specific policies are weighted and combined based on the currently desired task, which is dynamically inferred from the user's input. Our proposed adaptation approach involves two steps: i) determining policy likelihood and ii) scaling policy importance.
We determine the \textit{policy likelihood} of all mission-specific \acp{RMP} by evaluating how well they match the observed human input. The \textit{policy scaling} step then prioritizes policies with likely and non-conflicting features.
In the following, both steps are presented in detail.
A conceptual 2D example is shown in \Cref{fig:2D_example}.
\subsection{Policy Likelihood}
\label{ssec:likelihood}
\subsubsection{Human-desired Motion and Likelihood Formulation}
As outlined in \Cref{ssec:problem}, the human corrective input $\uH$ provides insights into their desired potential gradient $\rmpdPotHuman$.
Adopting a modeling approach for human-interactive trajectory deformation~\cite{2018-LosMal,2021-LosBajOma}, this relationship is formalized as:
\begin{align}
    -\rmpdPotHuman = 
    \twistHR
    +\matr{K}
    \uH,
\end{align}
where $\matr{K}\in\nR{\dimH\times\dimH}_{>0}$ is a scaling matrix.
\begin{remark}
    If $\uH$ is normalized, the matrix $\matr{K}$ represents the maximum permissible human velocity adjustment. 
    This intuitive interpretation simplifies the tuning process.
\end{remark}
We then define the likelihood $\likelihoodTotShort\in[0,1]$ of each mission-specific \ac{RMP} $i=\{1,\dots,N_{mis}\}$ as: \begin{align}
\likelihoodTotShort:=\likelihoodPost{\rmpdPotHuman}{\left(\rmpF^\mfldH_i,\rmpA_{i}^\mfldH\right)}\cdot\likelihood{\left(\rmpF^\mfldH_i,\rmpA_{i}^\mfldH\right)},
    \label{eq:policy_likelihood}
\end{align}
where the first factor represents the conditional likelihood of achieving the human-desired gradient with this particular policy, and the second term is the policy prior.

\subsubsection{Conditional Likelihood \texorpdfstring{$\likelihoodPost{\rmpdPotHuman}{\left(\rmpF^\mfldH_i,\rmpA_{i}^\mfldH\right)}$}{}}
The conditional likelihood quantifies the disparity between the human-desired and policy induced potential gradient.
Assuming that the human primarily indicates the desired direction of motion rather than the actual speed (see \Cref{ssec:problem}), the conditional likelihood is modeled based on the alignment angle between these two gradients. We therefore measure the cosine similarity \add{which lies in $[-1,1]$}, utilizing the inner product and norm induced by the policy metric\add{, and project it onto $[0,1]$}:
\add{
\begin{align}
    \text{cos\_sim} = 
    \frac{\innerProd{\rmpdPotHuman}{\rmpdPot_i^\mfldH}_{\rmpA_{i}^\mfldH}}
    {
    \norm{\rmpdPotHuman}_{\rmpA_{i}^\mfldH}
    \cdot\norm{\rmpdPot_i^\mfldH}_{\rmpA_{i}^\mfldH}
    }\label{eq:policy_likelihood_cos_sim}\\
    \likelihoodPost{\rmpdPotHuman}{\left(\rmpF^\mfldH_i,\rmpA_{i}^\mfldH\right)}
    = \frac{1}{2}\left(1 + \text{cos\_sim}\right)
    \label{eq:policy_likelihood_posterior}
\end{align}}

\subsubsection{Policy Prior \texorpdfstring{ $\likelihood{\left(\rmpF^\mfldH_i,\rmpA_{i}^\mfldH\right)}$}{}}
Unlike the conditional likelihood, the policy prior is not influenced by the human-desired direction of motion. Instead, it incorporates more intrinsic factors associated with the policy.
We take into account two aspects: i) the optimality of the robot's current state with respect to the policy's underlying task feature, and ii) the policy's metric which encodes its relative importance compared to all other polices.

We denote $\poseHROpt{i}$ as the optimal state of each task feature expressed in the human manifold $\mfldH$, i.e., $\poseHROpt{i}=\text{argmin}_{\poseHR}\ \rmpPot_i^\mfldH(\poseHR)$.
Inspired by the goal detection in~\cite{2013-DraSri}, we formulate the policy prior by considering the distance vector $\vect{d}_i=\poseHR - \poseHROpt{i}\in\nR{\dimH}$ between the current and optimal state. This essentially creates an artificial region of attraction for the different policies.

The policy metric should account for the directional and axis-specific weighting properties (see \Cref{ssec:rmp_planner}). To \add{do} so, we use the spectral decomposition of $\rmpA_{i}^\mfldH$ to independently consider each principal axis and its associated weight in the calculation of 
$\likelihood{\left(\rmpF^\mfldH_i,\rmpA_{i}^\mfldH\right)}\add{\in[0,1]}$:
\begin{align}
    d_{i,j} &= (1-\norm{\uH})\innerProd{\eigVec_{i,j}^\mfldH}{\vect{d}_i}\label{eq:policy_likelihood_prior_dist}\\
\likelihood{\left(\rmpF^\mfldH_i,\rmpA_{i}^\mfldH\right)} &=
    \sum_{j=1}^{\dimH} 
    \exp\left(-\frac{d_{i,j}^2}{2\eigVal_{i,j}}
    \right)
    \cdot
    \frac{\eigVal_{i,j}^\mfldH}{\sum_{k}\eigVal_{i,k}^\mfldH},
    \label{eq:policy_likelihood_prior}
\end{align}
where $\eigVec_{i,j}^\mfldH\in\nR{3}$ and $\eigVal_{i,j}^\mfldH\in\nR{}_{\geq0}$ are the $j$-th eigenvector and corresponding eigenvalue of $\rmpA_{i}^\mfldH$, respectively.
The exponential term in \Cref{eq:policy_likelihood_prior} maps the projected distance $d_{i,j}\in[0,\infty)$ into a likelihood in the range $[0,1]$, following the well established \textit{maximum entropy} methodology~\cite{2008-ZieMaaBag}.
As such, \Cref{eq:policy_likelihood_prior} is a weighted average over the main policy directions, exhibiting increased sensitivity towards dominant directions characterized by high metric eigenvalues.

The projected distance $d_{i,j}\in[0,\infty)$ in \Cref{eq:policy_likelihood_prior_dist} is adjusted based on the amount of human input $\uH$. 
This ensures that the operator maintains full control over the system's motion, preventing the robot from becoming trapped in a policy's region of attraction.

\begin{remark}
The heightened sensitivity, as well as dividing the distance $d_{i,j}$ by the eigenvalue in \Cref{eq:policy_likelihood_prior} are deliberate design decisions. Firstly, directions emphasized in the construction of the metric $\rmpA_i^\mfldH$ tend to have a more significant impact. Secondly, high-value metrics generally increase the policy prior due to smaller $d_{i,j}$ compared to low-value ones.
The former ensures handling of intra-policy directional weighting, while the latter considers inter-policy relative weighting. 
\end{remark}

\subsection{Policy Scaling}
\label{ssec:scaling}
\add{Interpreting} the proposed task parametrization as a linear combination of \acp{RMP}, the current task is uniquely defined by the metric scaling values $\rmpScalei_i$ in \Cref{eq:task_motion}.
These values directly govern the influence of each \ac{RMP} and its associated task feature on the robot's trajectory.
At its core, task adaptation thus involves the adjustment of these coefficients.

The approach, outlined in \Cref{alg:policy_scaling}, processes policies in decreasing order of likelihood, i.e. giving preference to likely policies. Policies controlling task features that are independent of the current summed-up policy increase in importance. Conversely, policies aligned with the current overall policy are reduced in importance, as other higher-likelihood policies are already controlling the same task features. The individual steps outlined in the algorithm are explained in detail next.

\begin{remark}
Our approach differs from existing literature which executes only the most likely policy ~\cite{2017-JavAdmPel,2013-DraSri,2019-KhoBil}.
This limits the ability to track independent/orthogonal task features (e.g., maintaining end-effector orientation while moving towards a goal position). As our approach leverages \acp{RMP}, we can naturally combine multiple policies into a single coherent one. \end{remark}

\begin{algorithm}
    \KwIn{$N_{mis}$ mission-specific policies $\left(\rmpF_i^\mfldH,\rmpA_i^\mfldH\right)$ and corresponding likelihoods $P_i$}
    \KwOut{policy scales $\rmpScalei_i$}

    \BlankLine
    Sort policies in decreasing order of likelihood\;

    \add{Initialize cumulative policy \begin{small}$\left(\rmpF_{P_i>P_j}^\mfldH,\rmpA_{P_i>P_j}^\mfldH\right)=\left(\vect{0},\matr{0}\right)$\end{small}\;}
    \BlankLine
    \ForEach{policy $\left(\rmpF_j^\mfldH,\rmpA_j^\mfldH\right)$ in sorted list}{
        \tcp{1.) alignment computation}
         \add{$\rmpScaleAngle = \frac{ \innerProd{\rmpdPot_{P_i>P_j}^\mfldH}{\rmpdPot_j^\mfldH}_{ 
         \rmpA_{P_i>P_j}^\mfldH}}
         {
            \norm{\rmpdPot_{P_i>P_j}^\mfldH}_{\rmpA_{P_i>P_j}^\mfldH}\cdot
            \norm{\rmpdPot_j^\mfldH}_{\rmpA_{P_i>P_j}^\mfldH}
         }$}
         \;\label{arg:policy_scaling_cos_sim}

         \BlankLine
        \tcp{2.) scale update}
        \If{$\rmpScaleAngle = 0$}{
            $\alpha_j \mathrel{+}= \rmpScaleDelta$\;
        }
        \Else{
            $\alpha_j \mathrel{-}= \rmpScaleDelta$\;
        }
        Clamp $\alpha_j$ to be in $[0,1]$\;

        \BlankLine    
        \tcp{3.) cumulative policy update}
        \add{
        $\left(\rmpF_{P_i>P_j}^\mfldH,\rmpA_{P_i>P_j}^\mfldH\right) \mathrel{+}= \left(\rmpF_{j}^\mfldH,\alpha_j\rmpA_{j}^\mfldH\right)$ 
        }\; \label{alg:policy_scaling_rmp_sum}

    }

    \BlankLine
    \caption{Policy Scaling}
    \label{alg:policy_scaling}
\end{algorithm}

\subsubsection{Alignment Computation}
\label{sssec:alignment_comp}
\add{We use cosine similarity between policy gradients to assess if their task features conflict.}
Specifically, as indicated in \Cref{arg:policy_scaling_cos_sim}, we use the \add{cosine similarity} between the iterator policy's objective gradient $\rmpdPot_j^\mfldH$ and that of all higher-likelihood policies $\rmpdPot_{P_i>P_j}^\mfldH$ (see \Cref{sssec:summed_policy_update}).
Similar to \Cref{eq:policy_likelihood_cos_sim}, we utilize the inner product induced by the associated metric \add{$\rmpA_{P_i>P_j}^\mfldH$}, adhering to the principles of Riemannian geometry~\cite{2018-RatIssKap}.

\subsubsection{Scale Update}
Depending on the similarity returned by the alignment computation, the metric scales $\rmpScalei_i$ are \add{adjusted}. Intuitively, task feature gradients are declared non-conflicting when the alignment angle $\rmpScaleAngle$ is exactly $\SI{90}{\degree}$ and conflicting otherwise. The actual adjustment of the scales is done gradually to ensure continuity, as required by stability guarantees for \acp{RMP}~\cite{2018-CheMukIss}. Fine-tuning the adjustment step $\rmpScaleDelta$ affects the dynamics of task adaptation. Large values enhance responsiveness to human inputs, while smaller values reduce sensitivity to erroneous operator commands~\cite{2021-LosBajOma}, resulting in overall smoother transitions between policies.

\subsubsection{\add{Cumulative} Policy Update}
\label{sssec:summed_policy_update}
The cumulative policy \add{$\left(\rmpF_{P_i>P_j}^\mfldH,\rmpA_{P_i>P_j}^\mfldH\right)$} combines all higher-likelihood polices with scaled metrics. 
In this context, the $\mathrel{+}=$ operator in \Cref{alg:policy_scaling_rmp_sum} implements the \add{addition of two \acp{RMP}} according to \Cref{eq:rmp_sum}.

\subsection{Stability \add{and Robustness} Considerations}
\label{ssec:stability}
In this section, we briefly comment on several stability and robustness-related aspects of the developed methodology. 

Firstly, the motion control framework is bounded-input-bounded-output stable since it is driven by asymptotically stable motion policies, and no energy is introduced from human manual control.
Secondly, if the human perfectly demonstrates the desired task, the task adaptation converges to the correct scaling values $\rmpScalei_i$. 
Finally, once the task adaptation has converged, the robot motion $\uR$ optimizes the human reward $\rH$ in a globally asymptotically stable manner.
Together, these properties guarantee the stability and functionality of the overall system. \add{The detailed mathematical analysis of \add{these} properties is provided in the \hyperref[app:appendix]{Appendix}.}

\add{
In uncertain environments, discrepancies between the safety-critical/mission-specific \ac{RMP} definitions and real-world conditions may arise.
However, task adaptation maintains the stated stability and convergence properties by relying solely on the instantaneous potential gradient of predefined task features, regardless of the precision and accuracy of these features.

\begin{remark}
    Failure to align task features with the real-world environment can still hinder mission success, despite robust task adaptation.
    While we leave specific investigations into robustifying features for future work, the modular nature of our methodology inherently permits real-time adjustment or addition of individual motion policies based on evolving environmental understanding, without modifying the motion planning and task adaptation framework.
\end{remark}
} 
\section{Experimental Setup}
\label{sec:exp_setup}
\begin{figure}
    \centering
    \includegraphics[width=\linewidth]{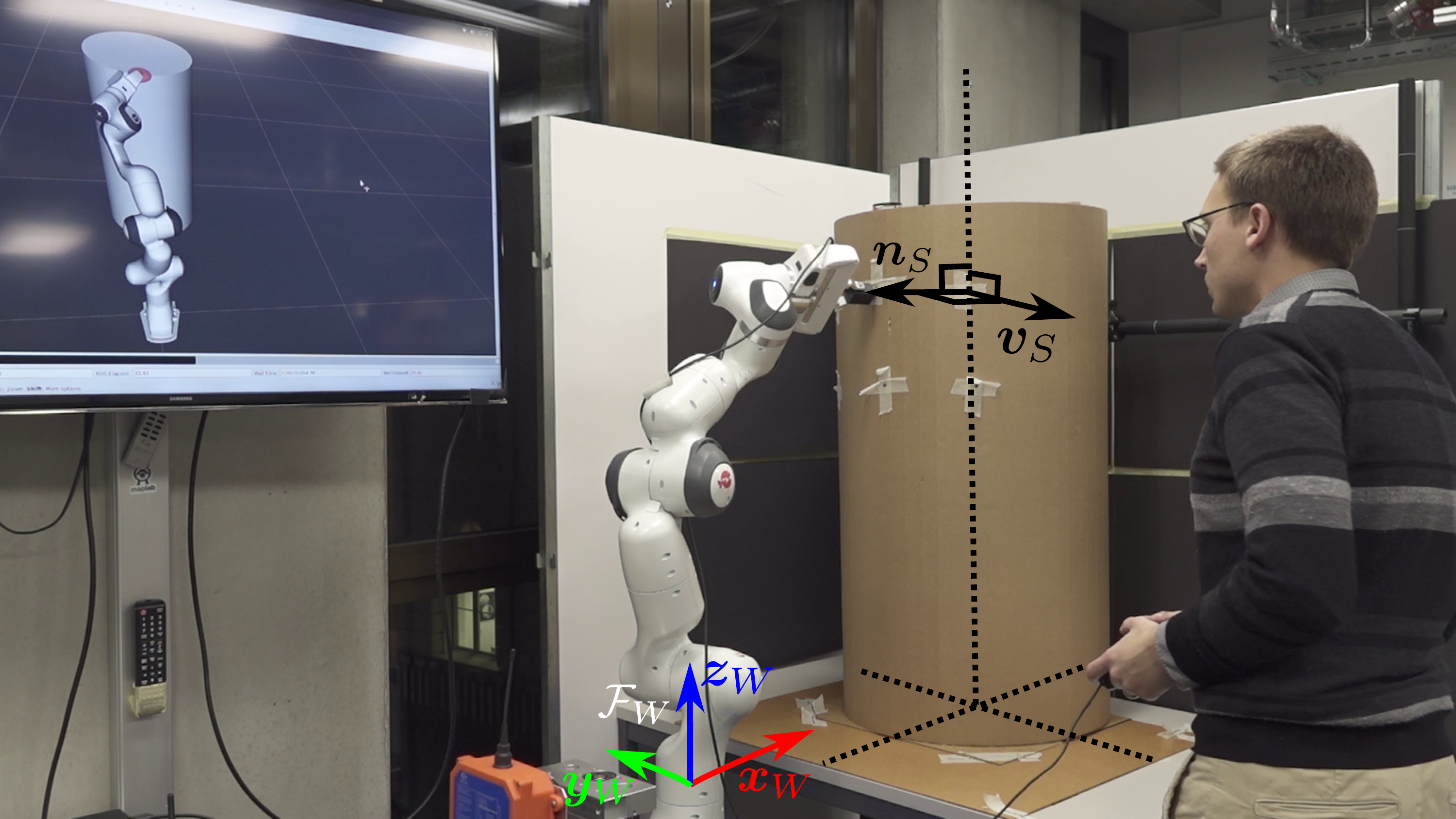}
    \caption{Experimental setup: cylindrical surface with inspection targets for the Franka Panda arm, user input via Gamepad, and visual feedback in RViz. The inertial world frame $\frameW$, as well as the instantaneous surface normal $\bm{n}_S$ and tangent $\bm{v}_S$ are indicated.}\label{fig:demo_view}
\end{figure}

In this section, we outline the experimental setup and evaluation scenario for the quantitative analysis presented in \Cref{sec:results}.
The supplementary video material\footnote{\url{https://youtu.be/E5SlSKx7pQk}} showcases the setting, as well as demonstrating additional applications. 

\subsection{General Mission Description}
\label{ssec:mission_description}
Our objective is to replicate a non-destructive testing application~\cite{2018-Ang}, using a 7\ac{DoF} Franka Panda arm equipped with a prism-like end-effector imitating the sensor.
Inspired by applications in oil tank and wind turbine inspections, the experimental setup features a cylindrical object with six inspection targets on its surface (see \Cref{fig:demo_view}).

To ensure accurate measurements, the sensor must be positioned in close proximity to an inspection point-of-interest, while maintaining alignment with the surface normal $\bm{n}_S$. Additionally, a minimum distance from the object should be ensured for safe operation when not recording data.
To interact with the robot, user commands $\uH$ are entered using a Gamepad (Logitech F310), providing 4\ac{DoF} inputs with continuous joysticks. 
As the application is intricately tied to the object surface, we define the human input manifold to encompass 2D motion on the surface along $\bm{z}_W$ and $\bm{v}_S$, and end-effector rotation around $\vect{n}_S$.
Situational awareness is ensured by allowing the user to maintain line-of-sight with the physical robot or through a digital twin of the setup via a 3D rendered visualization, as depicted in \Cref{fig:demo_view}.

\subsection{Task Feature Formulation}
\label{sssec:task_feature_formulation}
For this application, we define the following motion policies and matching manifolds to capture relevant task features:
\begin{description}
    \item[Normal-Keeping \ac{RMP}]
    \textit{Safety-critical} policy that ensures sensor alignment with $\bm{n}_S$. Acting on the $\mfldX_1=\SO{3}$ manifold of end-effector orientations.

    \item[Distance-Keeping \ac{RMP}]
    \textit{Safety-critical} policy that ensures safe distance. Acting on the $\mfldX_2=\nR{}$ manifold along $\bm{n}_S$.

    \item[Inspection Position \acp{RMP}]
    \textit{Mission-specific} policies $\rmpF_{pos,i},\ i\in\{1,\cdots,6\}$ that pull the end-effector towards the respective inspection target.
    Acting on the $3$-dimensional manifold $\mfldX_3=\nR{}\times\mathcal{S}^1\times\nR{}$ encompassing the object surface and distance to it.

    \item[Inspection Rotation \acp{RMP}]
    \textit{Mission-specific} policies $\rmpF_{rot,j},\ j\in\{1,2\}$ that rotate the sensor around $\bm{n}_S$ into a horizontal or vertical orientation (see below).
    Acting on the $\mfldX_1$ manifold of end-effector orientations.\end{description}
The specific formulation of these policies is omitted here for brevity but follows the generic target attractor outlined in~\cite{2018-RatIssKap}.

\begin{figure}
    \centering
    \includegraphics[width=0.9\linewidth]{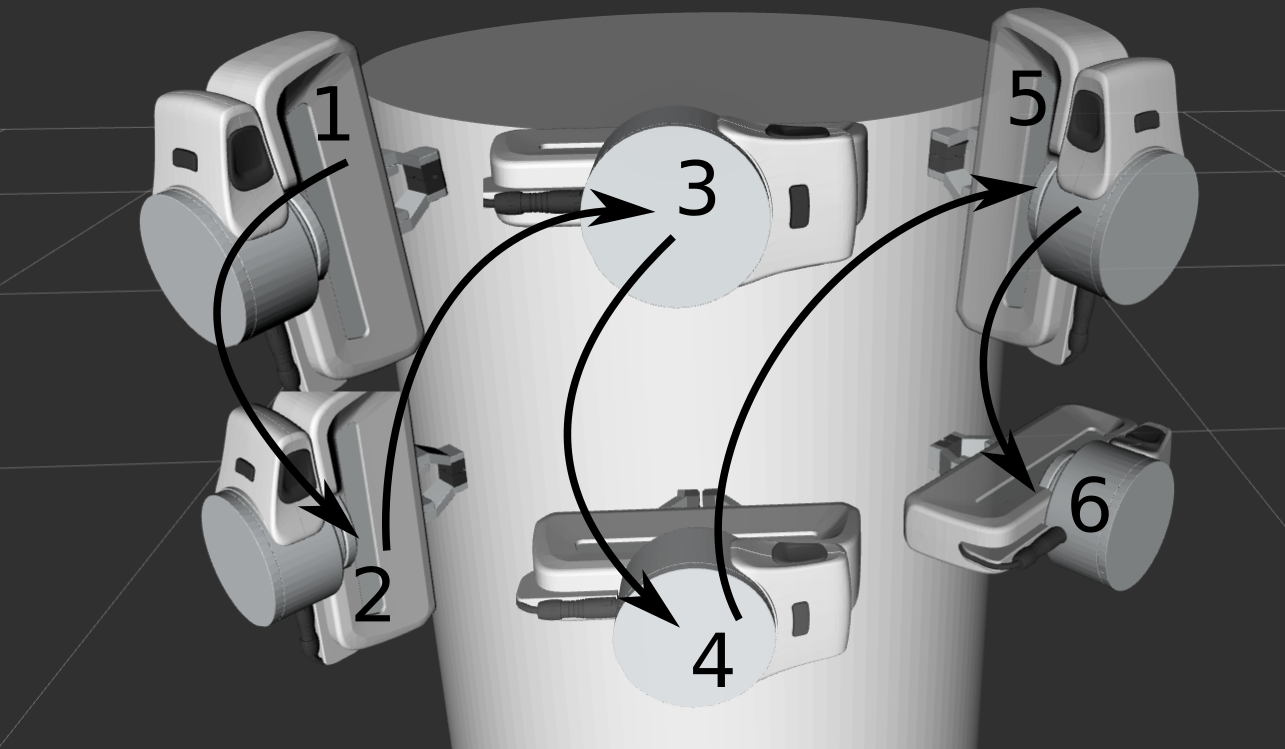}
    \caption{Desired sequence of inspection targets and sensor rotations.}\label{fig:goal_order}
\end{figure}

\subsection{Evaluation Scenario}
\label{ssec:eval_scenario}
The operator receives visual instructions through the 3D visualization, indicating the next position to inspect and a desired tool orientation according to \Cref{fig:goal_order}.
The task is to move the end-effector within $\pm\SI{5}{\centi\meter}$ and $\pm\SI{3}{\degree}$ of the indicated pose, as fast as possible.

We evaluate our methodology both independently and in comparison with the state-of-the-art task adaptation method in~\cite{2019-KhoBil}, hereafter referred to as \textit{Sota}.
\add{This work is deemed representative, as it addresses industrial \ac{HRI}, allows parametrization of generic tasks, and features a comparable task adaptation setup.}
By \add{additionally} assessing our method against \add{two} different forms of purely manual robot control we gain insight into how our exclusively autonomous framework compares in input efficiency.
\add{
To ensure optimal performance and fairness in comparison, all experiments are conducted by a group of $N=7$ robotic researchers, generally experienced in operating robotic systems but not familiar or trained on this experimental setup.}
\add{All policies and adaptation laws run at \SI{100}{\hertz}}.

The different methodologies are briefly discussed below:
\subsubsection{Ours}
This method adopts the dynamic task adaptation and motion control framework proposed in this paper. 
Following \Cref{ssec:mission_description}, the human input manifold allows motion on the object surface in vertical and tangential directions, as well as end-effector rotation around the surface normal. The robot motion is driven by combining \acp{RMP} to maintain normal orientation, ensure safety distance, and move towards inspection targets.

\subsubsection{2D Manual Control}
This method adopts standard rate control teleoperation, where the joystick controls the robot's velocity.
The human input is mapped to the curved surface of the cylinder, which relieves the operators from exact control along $\bm{n}_S$.
Unlike \textit{Ours} however, the robot only moves in response to non-zero user inputs and does not track inspection poses.
The task adaptation in \Cref{sec:method} is executed in the background for evaluation purposes.
Normal- and Distance-Keeping is still ensured by the corresponding \acp{RMP}.

\subsubsection{3D Manual Control}
Similar to the \textit{2D} approach, this method employs a rate control setup to manually control the robot's velocity. However, the operator is now responsible for also controlling the distance to the surface along $\bm{n}_S$.
Task adaptation data is again recorded for evaluation only. Robot autonomy is now limited to orientation of the end-effector in order to remain normal to the surface.

\subsubsection{Sota}
\label{sssec:sota}
The methodology outlined in~\cite{2019-KhoBil} represents tasks as 1st-order dynamical systems. This involves the definition of a function that maps the robot's current pose into a velocity that optimizes all corresponding features.
The task adaptation mechanism in~\cite{2019-KhoBil} compares the human-adjusted robot velocity with each task, updating the respective probability depending on how closely they match.
Note that robot motion control, task formulation, human interaction, and task adaptation are all designed in the same manifold, namely the robot configuration space $\SE{3}$.

\section{Quantitative Evaluation}
\label{sec:results}
Based on the setup presented in \Cref{sec:exp_setup}, we conduct a quantitative assessment of the proposed methodology.
The results and discussion address critical performance aspects, considering the methodology both independently and in comparison with the state-of-the-art task adaptation method in~\cite{2019-KhoBil}.

\subsection{Results}
\label{ssec:results}
We focus the evaluation of our method and comparison with \textit{Sota} on how quickly the respective adaptation converges to the human-desired task and the amount of user input required to achieve this convergence.
\add{
To emphasize, convergence refers to the time required for task adaptation (\Cref{sec:method} for \textit{Ours}, \textit{2D}, \textit{3D}; or~\cite{2019-KhoBil} for \textit{Sota}) to identify the correct target, and not completion of the task itself.}
The convergence times for inspection position and rotation of each task are reported in \Cref{fig:scale_convergence}, visualizing the distribution across the $N=7$ operators for each methodology.
\add{
    In essence, this represents the minimum interaction time required before the robot can autonomously complete the task, which correlates with the expressiveness of human input for the different methodologies.
}

In terms of user input, statistics for the total human effort \add{needed} to complete the mission are provided in \Cref{tab:statistics}. 
\Cref{fig:user_input} illustrates the accumulated operator commands along the individual joystick axes during the transition from task $2$ to $3$ (see \Cref{fig:goal_order}).  
This transition offers detailed insights into fundamental differences between the four methods and, specifically, between \textit{Ours} and \textit{Sota}. 

For completeness, we also report statistics on the task completion time and accuracy, represented as the minimum translational and rotational error, in \Cref{tab:statistics}.
\add{Although these metrics are subject to the influence of the robot and tuning of the individual \acp{RMP} or dynamic systems,
the table offers a high-level comprehensive comparison of autonomous (\textit{Ours}/\textit{Sota}) and manual (\textit{2D}/\textit{3D}) methods.
Indeed, the results confirm that pure manual control is generally unsuitable, resulting in increased user effort, longer task completion time and degraded accuracy. More fundamental insights into our methodology are discussed below.}

\begin{table}[ht]
    \centering
    \resizebox{\linewidth}{!}{
    \begin{tabular}{c||c|c|c|c}
        Method & Ours & 2D & 3D & Sota \\\hline
         \multirow{ 2}{*}{$\frac{1}{T}\int^T_0\norm{\uH(t)}^2dt$}
        &$\boldsymbol{\mu=41.15}$
        &$\mu=64.84$
        &$\mu=49.91$
        &$\mu=54.09$\\

         &$\sigma=11.35$
         &$\sigma=8.74$
         &$\boldsymbol{\sigma=8.33}$
         &$\sigma=15.71$\\\hline

        Task completion time
        &$\boldsymbol{\mu=11.19}$
        &$\mu=19.98$
        &$\mu=27.94$
        &$\mu=13.19$\\

        $[s]$
         &$\boldsymbol{\sigma=5.24}$
         &$\sigma=10.28$
         &$\sigma=14.95$
         &$\sigma=6.75$\\\hline

        Translation error
        &$\mu=0.89$
        &$\mu=6.74$
        &$\mu=9.87$
        &$\boldsymbol{\mu=0.57}$\\

        $[mm]$
        &$\boldsymbol{\sigma=0.30}$
        &$\sigma=3.88$
        &$\sigma=4.45$ 
        &$\sigma=0.40$\\\hline

        Rotation error
        &$\boldsymbol{\mu=0.19}$
        &$\mu=2.74$
        &$\mu=2.56$
        &$\mu=0.66$\\

        [\si{\degree}]
        & $\boldsymbol{\sigma=0.12}$
        & $\sigma=1.21$
        & $\sigma=1.34$
        & $\sigma=0.36$
    \end{tabular}}
    \caption{Mean $\mu$ and standard deviation $\sigma$ of different metrics across all $N=7$ operators.} 
    \label{tab:statistics}
\end{table}

\subsection{Discussion}
\label{ssec:results_discussion}

\begin{figure*}[ht!]
    \centering
    \includegraphics[width=\linewidth,height=6cm]{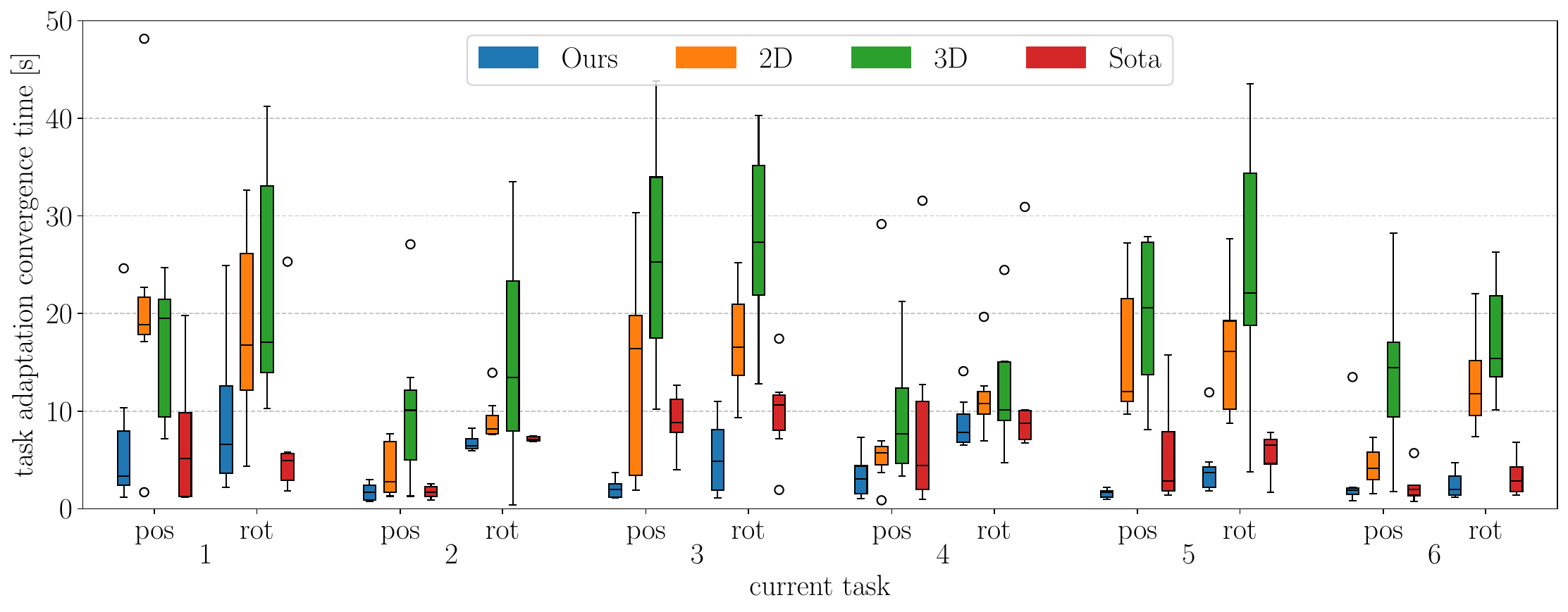}
    \caption{Task adaptation convergence time for inspection position (pos) and rotation (rot) part of each task across the $N=7$ operators.
    \add{Note that task adaptation is non-blocking, running in real-time at the same rate as robot motion control. This enables the robot to remain in motion and move towards the most likely target even prior to convergence.}
   }  
    \label{fig:scale_convergence}
\end{figure*}

\begin{figure}[htb]
    \centering
    \includegraphics[width=\linewidth,height=7cm]{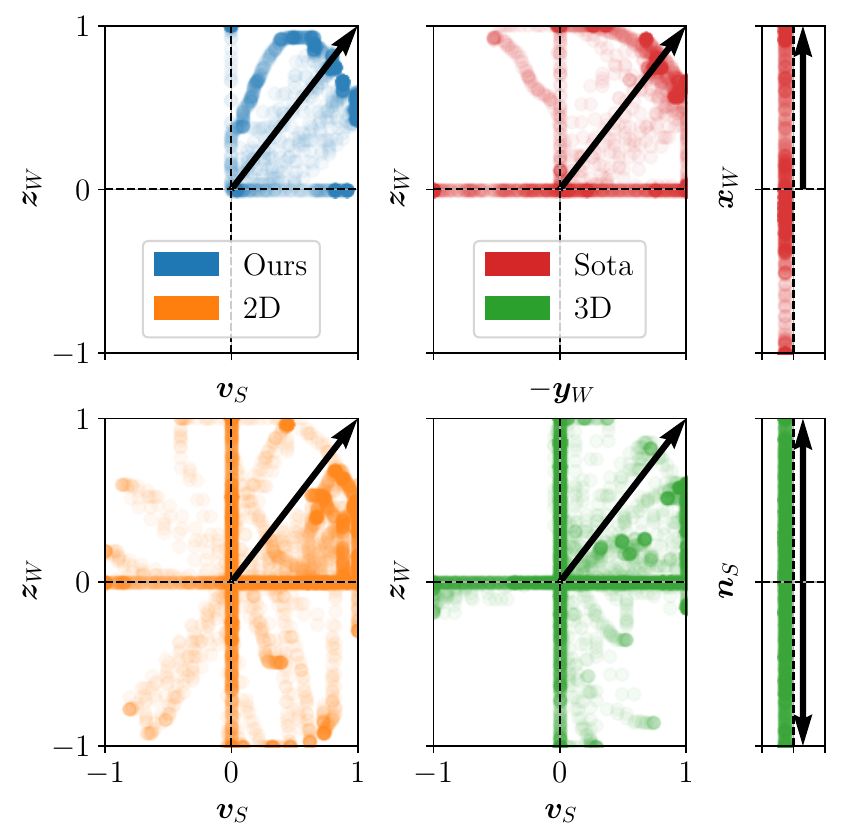}
    \caption{Translational joystick inputs for transition from task $2$ to $3$, accumulated over all $N=7$ operators. Dots correspond to inputs commanded by the users, with opacity corresponding to the occurrence frequency. 
    Labels indicate the mapped robot motion axis while black arrows represent the optimal commands for convergence of the respective task adaptation methodology. 
    Notably, the input manifolds $\mfldH$ for \textit{Ours} and \textit{2D} feature two translational control inputs, while the ones for \textit{3D} and \textit{Sota} feature three.
    }    
    \label{fig:user_input}
\end{figure}

A thorough investigation of the task adaptation convergence time and user inputs allows us to draw several conclusions.

As illustrated in \Cref{fig:scale_convergence}, dynamic task adaptation demonstrates significantly faster convergence to the correct task when no manual control is involved. 
This aligns with observations from \Cref{fig:user_input}, indicating that operator commands in \textit{Ours} are highly task oriented compared to \textit{2D} and \textit{3D}.
In our approach, all joystick commands are concentrated in the first quadrant, aligning with the optimal task direction and facilitating the rapid identification of the human-desired task.
Contrastingly, inputs using the two manual control methods are generally less informative for task adaptation. 
This discrepancy likely arises from the precision required to maneuver the robot under manual control, leading to numerous small-scale direction-changing corrective inputs.
This conclusion strongly supports the use of an \textit{exclusively autonomous} motion control framework.

When comparing with \textit{Sota}, their task adaptation achieves similar convergence rates as ours.
However, the required human input is much more intricate.
As the human input manifold is the cartesian configuration space $\SE{3}$, the motion needs to be carefully coordinated for the human to be able to correctly indicate the task along the curved surface.
This becomes particularly evident when considering the translational control inputs during the transition from task 2 to 3. 
In \textit{Ours}, where the input manifold $\mfldH$ encompasses the surface, convergence is achieved by a 2D input along $\bm{v}_S$ and $\bm{z}_W$, as indicated by the black arrows in \Cref{fig:user_input}.
In \textit{Sota}'s $\SE{3}$ manifold however, this necessitates coordinated motion along all three principal axes, $\bm{x}_W,\ -\bm{y}_W$ and $\bm{z}_W$. 
When moving along $-\bm{y}_W$, providing control input along $\bm{x}_W$, i.e., pushing the robot towards the surface, is crucial, since the task adaptation otherwise converges on task $5$ or $6$ due to the ambiguity in motion.
Many operators found this challenging, as evidenced by the strong corrective actions along $\bm{y}_W$ in \Cref{fig:user_input}. In fact, oral intervention from the experiment supervisor was often necessary, since the required correction along $\bm{x}_W$ was not self-evident.
Consequently, this transition exhibits the largest median convergence time for \textit{Sota} across all tasks.
This observation highlights the advantages of a \textit{purely local} motion planning framework, allowing us to leverage the inherent geometry of applications and design appropriate input manifolds.

Finally, we highlight key differences between our \textit{modular} approach and the full body motion planning used in \textit{Sota}. 
In \textit{Sota}, each task is expressed as a single $\SE{3}$ dynamical system, responsible for controlling every aspects of robot motion. This means \add{that} all desired behaviors, task completion, safety measures, and system constraints \add{have to be formalized} within a single function.
The design and debugging of these functions becomes notably challenging. 
Independent testing of individual components is often not feasible, and given that many aspects, e.g. safety considerations, are common across all tasks, redundant definitions are necessary. 
Indeed, experiments in~\cite{2019-KhoBil} required prior learning-from-demonstration for dynamical systems that did not follow the simple form of linear or circular motion.    
This reduces overall adaptability, as even small changes in tasks or the environment necessitate redesigning the entire dynamical system. 

On the contrary, modular motion planning, exemplified by \acp{RMP}, allows us to represent tasks as a combination of smaller, individually testable policies. Each component addresses a specific challenge and can be geometrically combined with others to realize more complex motion. Such modular approaches provide a more versatile toolbox for various tasks, streamlining the overall design process and enhancing usability. Using the approach presented in this paper, we enable the use of such a modular framework for \ac{HRI}.

\add{
\subsection{Scalability Considerations}
\label{ssec:scalability}

Comparing the scalability of our method with \textit{Sota}~\cite{2019-KhoBil}, we focus on complexity in industrial applications arising from non-intuitive task geometries and intricate task features.

The quantitative evaluation highlights the potential for enhanced intuitiveness through shaping the human input manifold $\mfldH$, resulting in superior user interaction compared to \textit{Sota}, or at worst, similar performance when selecting $\mfldH$ the same as theirs.

As task features increase in number and sophistication, designing a monolithic dynamical system for \textit{Sota} becomes increasingly challenging.
In contrast, our modular approach maintains simplicity and functional task adaptation by integrating independent and composable policies.
Challenges can arise when multiple task features have closely aligned potential gradients, which diminishes the expressiveness of human input and may prolong task adaption time.
Possible solutions are the inclusion of exaggerated robot motions to elicit more informative human inputs~\cite{2022-JonLos} or using augmented reality techniques~\cite{2021-MulMosCha}.} 

\section{Conclusion}
\label{sec:conclusion}
We introduce a motion control framework designed for industrial applications, streamlining dynamic task planning and adaptation to human operator commands. Our method eliminates the need for direct manual control, opting for exclusively autonomous motion planning without restricting human control authority over robot motion.
Task parametrization and adaptation leverage purely local and modular motion policies in the form of \aclp{RMP}. This enables a straightforward formulation of task features in their native manifold and provides a geometrically consistent way to combine them. 
To do so, we contribute a novel method to infer user intent directly in the form of a re-weighting of policy metrics, effectively exploiting the mathematical structure of \aclp{RMP}.

Quantitative experiments show that our method enables smooth and seamless \acl{HRI}. Moreover, we validate the potential of exclusively autonomous frameworks, showing that human manual control tends to increase the convergence time of task adaptation. In comparison to the state-of-the-art, we emphasize the advantages of exploiting the inherent geometry of tasks, simplifying the user interface, as well as the formulation of the tasks themselves.
\add{
In future work, we aim to enhance the robustness of our framework against inaccuracies in task features within uncertain environments. 
This entails refining \acp{RMP} in real-time based on evolving environmental understanding, as well as the identification of new task features on-the-fly.}

\appendix
\label{app:appendix}

\label{app:stability}
We provide the mathematical analysis for the stability properties outlined in \Cref{ssec:stability}.
Firstly, we discuss the stability of the generated motion reference $\uR$. 
Secondly, we analyze the extent to which the motion reference $\uR$ optimizes the human objective $\rH$.
Lastly, we delve into how the system converges to the human desired task.

\subsection{Stability}
\label{sssec:stability}
Referring to the stability analysis in~\cite{2018-CheMukIss}, \acp{RMP} of the form \Cref{eq:rmp_f} asymptotically drive the robot towards a steady-state $\poseX^\star$ that optimizes the underlying task feature, i.e. $\rmpdPot^\mfldH(\poseX^\star)=0$.
This applies to individual motion policies, as well as when applying the \textit{pull-back} \Cref{eq:rmp_pull} or addition operator \Cref{eq:rmp_sum}. 
The generated motion reference $\uR$ in \Cref{eq:task_motion} linearly combines stable motion controllers with $\rmpScalei_i\in[0,1]$.
Therefore, the absence of energy input from human manual control ensures that the motion control framework is bounded-input-bounded-output stable.
When the scales converge, i.e., $\dot{\rmpScalei}_i=0\ \forall i$, conditions from \cite{2018-CheMukIss} apply, establishing asymptotic stability of the system. 

\subsection{Optimality}
\label{sssec:optimality}
Closer examination of the robot motion reference $\uR$ reveals
a mission-specific potential gradient $\rmpdPot^\mfldC_{R,mis}$ of the form:
\begin{align}
    \rmpdPot^\mfldC_{R,mis}(\poseR) =
    \left(
    \sum_{i=1}^{N_{mis}} \rmpScalei_i\rmpA_i^\mfldC
    \right)^\dagger
    \left(
        \sum_{i=1}^{N_{mis}} \rmpScalei_i \rmpA_i^\mfldC \rmpdPot_i^\mfldC(\poseR)
    \right).
\end{align}

Given the stability properties outlined in the previous section, $\uR$ ultimately achieves a robot steady-state $\poseR^\star$ such that:
\begin{align}
      \vect{0} = \sum_{i=1}^{N_{mis}} \rmpScalei_i \rmpA_i^\mfldC \rmpdPot_i^\mfldC(\poseR^\star)
\end{align}
\begin{remark}
    We disregard safety-critical policies in this discussion, since their impact is negligible when the robot is not at risk, i.e. distant from obstacles or joint/workspace limits.
\end{remark}

When considering the human reward gradient expressed in configuration space:
\begin{align}
    \left(\frac{d \rH}{d \poseHR}\right)^\mfldC
    =
    \sum_{i=1}^{N_{mis}} \rmpScalei_i \rmpdPot_i^\mfldC(\poseR),
\end{align}
we see that the steady-state $\poseR^\star$ optimizes the human reward when the mission-specific metrics are equal for all $i$,
such as in the case of unknown task hierarchy.
More importantly, $\uR$ ultimately optimizes $\rH$ when the scales have converged on a single policy $i^\star$ that represent the current task, i.e., $\rmpScalei_{i^\star} = 1,\ \rmpScalei_i=0\ \forall i\neq i^\star$. 

\subsection{Convergence to Demonstration}
\label{sssec:demo}
Assume the operator provides a perfect demonstration for the desired task, given as $\rmpPot^\mfldH_{i^\star}$, with maximum effort. In other words, $\norm{\uH}=1$ and:
\begin{align}
    \frac{\rmpdPotHuman}{\norm{\rmpdPotHuman}} 
    = 
    \frac{\rmpdPot^\mfldH_{i^\star}}{\rmpdPot^\mfldH_{i^\star}}, 
\end{align}

For the conditional \Cref{eq:policy_likelihood_posterior} and prior \Cref{eq:policy_likelihood_prior}, it then holds that
\begin{align}
\likelihoodPost{\rmpdPotHuman}{\left(\rmpF^\mfldH_{i^\star},\rmpA_{i^\star}^\mfldH\right)} 
&= \max_i \likelihoodPost{\rmpdPotHuman}{\left(\rmpF^\mfldH_{i},\rmpA_{i}^\mfldH\right)} 
\\
\likelihood{\left(\rmpF^\mfldH_i,\rmpA_{i}^\mfldH\right)} &= 1\ \forall i, 
\end{align}
which in turn implies that 
\begin{align}
    \dot{\rmpScalei}_{i^\star} > 0 && \dot{\rmpScalei}_{i} < 0\ \forall i\neq i^\star. 
\end{align}
Eventually, the dynamic task adaptation converges to $\rmpScalei_{i^\star} = 1,\ \rmpScalei_i=0\ \forall i\neq i^\star$, optimizing the desired task feature (see \Cref{sssec:optimality}) in a globally asymptotically stable manner (see \Cref{sssec:stability}). 

\bibliographystyle{unsrtnat}

\bibliography{bibliography/intro, 
                bibliography/rmp,
                bibliography/shared_autonomy}

\end{document}